\documentclass{article}

\PassOptionsToPackage{numbers, compress}{natbib}

\usepackage[main, final]{neurips_2026}

\usepackage[utf8]{inputenc} 
\usepackage[T1]{fontenc}    
\usepackage{hyperref}       
\usepackage{url}            
\usepackage{booktabs}       
\usepackage{amsfonts}       
\usepackage{nicefrac}       
\usepackage{microtype}      
\usepackage[table]{xcolor}         

\usepackage{multirow}
\usepackage[normalem]{ulem}
\usepackage{algorithm}
\usepackage{algorithmic}
\usepackage{amsmath}
\usepackage{tcolorbox}
\usepackage{svg}
\tcbuselibrary{skins}
\title{FlowErase-RL: Rethinking Concept Erasure as Reward Optimization in Flow Matching Models}

\author{%
Yi Sun$^{1}$ \quad Zhiqi Zhang$^{3}$ \quad Xinhao Zhong$^1$ \quad Yimin Zhou$^{2}$ \quad Shuoyang Sun$^1$ \\
\bfseries  Bin Chen$^{1,4}$\thanks{Corresponding Author.} \quad
 Shu-Tao Xia$^2$ \quad Ke Xu$^5$\\
$^1$Harbin Institute of Technology, Shenzhen \\ 
$^2$Tsinghua Shenzhen International Graduate School, Tsinghua University \\
$^3$Jilin University \\
$^4$Peng Cheng Laboratory \\
$^5$Department of Computer Science and Technology, Tsinghua University \\
}

\begin{document}

\maketitle

\begin{abstract}

Recent advances in flow matching models have significantly improved text-to-image generation quality, but also introduce growing safety risks due to the generation of harmful or undesirable content. Existing concept erasure methods are either inference-time interventions with limited effectiveness or rely on supervised fine-tuning (SFT), which requires precisely aligned data and struggles with scalability and multi-concept settings. In this paper, we propose \emph{FlowErase-RL}, the first GRPO-based framework for concept erasure in flow matching models. We reformulate concept erasure as a reward optimization problem and introduce a \textbf{dynamic dual-path reward mechanism} that jointly optimizes (i) a Concept Erasure (CE) reward to suppress target concepts and (ii) a Non-target Space (NS) reward to preserve generative fidelity. The two reward paths are adaptively balanced during training via a performance-driven switching strategy, enabling stable optimization without explicit supervision. Extensive experiments on nudity, object, and artistic style erasure demonstrate that our method achieves state-of-the-art erasure performance while maintaining strong image quality and semantic alignment. Moreover, it exhibits robust resistance to adversarial attacks and scales effectively to multi-concept scenarios. Our results establish a new paradigm for safe and controllable generation in flow matching models.

\end{abstract}    
\section{Introduction}
\label{sec:introduction}

In recent years, attributable to the uncurated nature of Internet-sourced training data~\cite{milmo2023ai}, the proliferation of inappropriate content generated by text-to-image (T2I) models~\cite{Dhariwal2021DiffusionMB,Ho2022ClassifierFreeDG,Ho2020DenoisingDP,Nichol2021GLIDETP,Rombach2021HighResolutionIS,Saharia2022PhotorealisticTD} has become a pressing concern. While the capability to synthesize high-quality images from textual prompts has advanced significantly, from the advent of DALL-E 2~\cite{Rombach_2022_CVPR} and Stable Diffusion (SD)~\cite{Rombach2021HighResolutionIS} to the beefed-up Flux~\cite{flux2024}, the risk of producing harmful outputs~\cite{10.1145/3600211.3604681,roose2022ai,setty2023ai,Ed2024ai} necessitates effective countermeasures. Complete data removal followed by model retraining~\cite{Nichol2021GLIDETP,StableDiffusion_2022,Schramowski2022SafeLD} represents a direct approach, yet its exorbitant computational overhead, inefficiency, and susceptibility to performance degradation~\cite{oconnor2022stable} render it impractical. In contrast, concept erasure (CE) offers a compelling alternative, enabling the suppression of specific target concepts through lightweight interventions while safeguarding the model's generative proficiency.

Existing CE methods have been developed primarily for the SD paradigm, which relies on the DDPM~\cite{Ho2020DenoisingDP} and DDIM~\cite{Song2020DenoisingDI} sampling procedures coupled with the U-Net backbone. In contrast, the Flux series departs substantially from this design. Flux adopts flow matching~\cite{Lipman2022FlowMF} as its core generative mechanism and employs a transformer-based architecture instead of the U-Net structure. Furthermore, Flux integrates an additional text encoder, Google T5, and utilizes Rotary Position Embedding (RoPE) for both pixel and textual representations. These architectural distinctions render prior CE techniques ineffective when applied to the Flux framework. Existing approaches, whether training-free methods~\cite{Wang2024PreciseFA,Chavhan2024ConceptPruneCE,Gandikota2023UnifiedCE} that merely intervene at inference time or training-based methods~\cite{Chen2025TRCETR,Lu2024MACEMC,Cywinski2025SAeUronIC,Kim2024RACERA,Li2024SafeGenMS, zhong2025closing} tailored to the SD pipeline, fail to generalize to this modern architecture. The inability to transfer concept erasure techniques from SD to Flux thus constitutes a critical gap that this work seeks to bridge.

To address this issue, we propose the first concept erasure method based on GRPO~\cite{Shao2024DeepSeekMathPT,Liu2025FlowGRPOTF} called \emph{FlowErase-RL}. By formulating target concepts as a CE reward function, FlowErase-RL achieves effective erasure performance. To further enhance erasure performance while preserving generation capability, we design a dynamic dual-path reward mechanism. The verifiable CE Reward guides the model to move away from the target concept during image synthesis, while the Non-target Space (NS) Reward encourages the model to generate high-quality images for prompts unrelated to the target concept. During training, the CE reward employs prompt pairs with or without target concepts and the NS reward takes a retain set as input. The two reward paths are automatically switched per epoch with their strength and weighting dynamically adjusted based on the observed erasure performance. Extensive experiments on multiple target concepts demonstrate that this dual-objective formulation simultaneously coordinates safety and utility within a unified reward structure, allowing FlowErase-RL to achieve superior erasure effectiveness while maintaining high generative quality. It outperforms existing state-of-the-art methods on flow matching models and also shows excellent erasure capability on other T2I model architectures. In summary, we make the following contributions:

\begin{itemize}
    \item We propose the first concept erasure method based on GRPO for flow matching models, achieving effective erasure while maintaining excellent generative capabilities. 
    \item We introduce a dynamic dual-path reward mechanism that adaptively switches reward functions during training, striking an effective balance between efficient concept erasure and the preservation of generative capabilities.
    \item Extensive experimental results based on flow matching models demonstrate that our method achieves or approaches SOTA performance on major erasure tasks.
\end{itemize}
\section{Related works}
\label{sec:related}

\subsection{Flow matching}

Flow matching~\cite{Lipman2022FlowMF} has become a powerful alternative to DDPM~\cite{Ho2020DenoisingDP} for generative modeling. While diffusion models learn to reverse a noise-adding process, flow matching directly learns the velocity field of an ODE that transports samples from noise to data. Rectified Flow~\cite{Liu2022FlowSA} proposes learning straight-line trajectories between noise and data pairs, enabling efficient single-step generation. Flux~\cite{labs2025flux1kontextflowmatching,flux2024} and Stable Diffusion 3~\cite{Esser2024ScalingRF} adopt flow matching with transformer architectures, achieving high image quality. Subsequent works have explored downstream tasks as well. Despite being efficient and high-quality, current flow matching generative models could still produce sensitive content and lack safety defense methods.

\subsection{Concept erasure methods in T2I models}

Existing CE methods~\cite{Rando2022RedTeamingTS,dalle3systemcard,Nichol2021GLIDETP,StableDiffusion_2022} can be categorized into two groups: training-free and training-based methods. Training-based methods typically involve fine-tuning pre-trained models to remove specific concepts while training-free methods often relay on causal intervention and masks to modify model behavior without fine-tuning during inference.  

\noindent\textbf{training-free methods} Unified Concept Editing (UCE)~\cite{Gandikota2023UnifiedCE} introduces closed-form solutions to edit the cross attention weights, changing the key and value matrices of specific text embeddings containing target concept, while retaining the matrix weights of unrelated concepts. Adaptive Value Decomposer (AdaVD)~\cite{Wang2024PreciseFA} disentangles target semantics from the cross-attention layers at each denoising timestep, dynamically controls the erasure intensity through an adaptive token-wise shift mechanism. ActErase~\cite{Sun2026ActEraseAT} try to identify activation difference regions via prompt-pairs and dynamically replace input activations during inference passes. Differential Vector Erasure (DVE)~\cite{Zhang2026DifferentialVE} constructs a differential vector field between target and anchor concepts and applies projection-based selective correction during inference to suppress undesirable semantics.

\noindent\textbf{training-based methods} Erased Stable Diffusion (ESD)~\cite{Gandikota2023ErasingCF} fine-tunes the latent diffusion model (LDM) by aligning the noise predictions of target and non-target concepts and guides this optimization via classifier-free guidance. Forget-Me-Not (FMN)~\cite{Zhang2023ForgetMeNotLT} focuses on fine-tuning the cross-attention maps to steer the generation of target concepts towards unrelated concepts. Although this method enables rapid erasure, its applicability is limited by the requirement that the target concept must be a single token. EraseAnything~\cite{gao2025eraseanything} introduces a Bi-level optimization approach that fine-tunes LoRA modules through a combination of attention‑map regularization and reversed contrastive learning, additionally enabling the migration of concept‑erasure techniques originally developed for diffusion models to the Flux framework. 
\section{Preliminaries}
\label{sec:preliminaries}

\subsection{Flow Matching Models} 
\label{subsec:fm}

Flow Matching~\cite{Lipman2022FlowMF} provides a simulation-free approach to training Continuous Normalizing Flows by directly regressing a neural network to a target velocity field. Since the marginal velocity is intractable, Conditional Flow Matching introduces a conditioning variable $\mathbf{x}_1 \sim p_{\text{data}}$ and defines a conditional probability path. For the straight-line interpolation $\mathbf{x}_t = (1-t)\mathbf{x}_0 + t\mathbf{x}_1$ with $\mathbf{x}_0 \sim \mathcal{N}(\mathbf{0}, \mathbf{I})$, the tractable CFM objective becomes Equation~\ref{eq:l_cfm}.

\begin{equation}
    \mathcal{L}_{\text{CFM}}(\theta) = \mathbb{E}_{t, \mathbf{x}_0, \mathbf{x}_1} \left[\left\| v_t^{\theta}(\mathbf{x}_t) - (\mathbf{x}_1 - \mathbf{x}_0) \right\|^2\right].
    \label{eq:l_cfm}
\end{equation}

For text-to-image generation, FLUX~\cite{flux2024} adapts this to the conditional setting with text prompt $\mathbf{c}$. The model predicts velocity $v_t^\theta(\mathbf{x}_t, \mathbf{c})$ conditioned on timestep $t$ and text embedding $\mathbf{c}$. The $\mathcal{L}_{\text{FLUX}}(\theta)$ can be represented by Equation~\ref{eq:l_flux}. 

\begin{equation}
    \mathcal{L}_{\text{FLUX}}(\theta) = \mathbb{E}_{t, \mathbf{x}_0, \mathbf{x}_1, \mathbf{c}} \left[\left\| v_t^{\theta}\left((1-t)\mathbf{x}_0 + t\mathbf{x}_1, \mathbf{c}\right) - (\mathbf{x}_1 - \mathbf{x}_0) \right\|^2\right],
    \label{eq:l_flux}
\end{equation}

where $\mathbf{x}_1$ is the latent encoding of an image. FLUX achieves efficient few-step sampling while maintaining high generation quality through a transformer-based architecture and guidance distillation.

\subsection{Group Relative Policy Optimization (GRPO)} 

GRPO \cite{Shao2024DeepSeekMathPT} eliminates the critic network by estimating advantages through group-based relative rewards. Given query $\mathbf{q}$ and $G$ responses $\{\mathbf{o}_i\}_{i=1}^G$ sampled from policy $\pi_\theta$, the advantage is normalized as $\hat{A}_i = (r_i - \bar{r}) / \sigma_r$ where $\bar{r}$ and $\sigma_r$ denote the mean and standard deviation of group rewards. The clipped surrogate objective with KL penalty is Equation~\ref{eq:l_grpo}.
\begin{equation}
    \mathcal{L}_{\text{GRPO}} = \mathbb{E}_{\mathbf{q}, \{\mathbf{o}_i\}} \left[\frac{1}{G} \sum_{i=1}^{G} \min\left(\rho_i \hat{A}_i, \text{clip}(\rho_i, 1-\epsilon, 1+\epsilon) \hat{A}_i\right) - \beta \mathcal{D}_{\text{KL}}\right],
    \label{eq:l_grpo}
\end{equation}
where $\rho_i = \frac{\pi_\theta(\mathbf{o}_i \mid \mathbf{q})}{\pi_{\text{old}}(\mathbf{o}_i \mid \mathbf{q})}$ and $\mathcal{D}_{\text{KL}} = \mathbb{D}_{\text{KL}}(\pi_\theta \| \pi_{\text{ref}})$.

Flow-GRPO~\cite{Liu2025FlowGRPOTF} extends this to flow-based models by treating velocity prediction as a sequential policy. For text prompt $\mathbf{c}$, it samples $G$ trajectories following the flow ODE and assigns the same group-normalized advantage $\hat{A}_i$, computed from the final reward, to all per-timestep velocities along each trajectory. This enables direct preference optimization of flow trajectories without reward model distillation.

\section{Method}
\label{sec:method}

\begin{figure*}[t]
  \centering
  \resizebox{\linewidth}{!}{
  \includegraphics{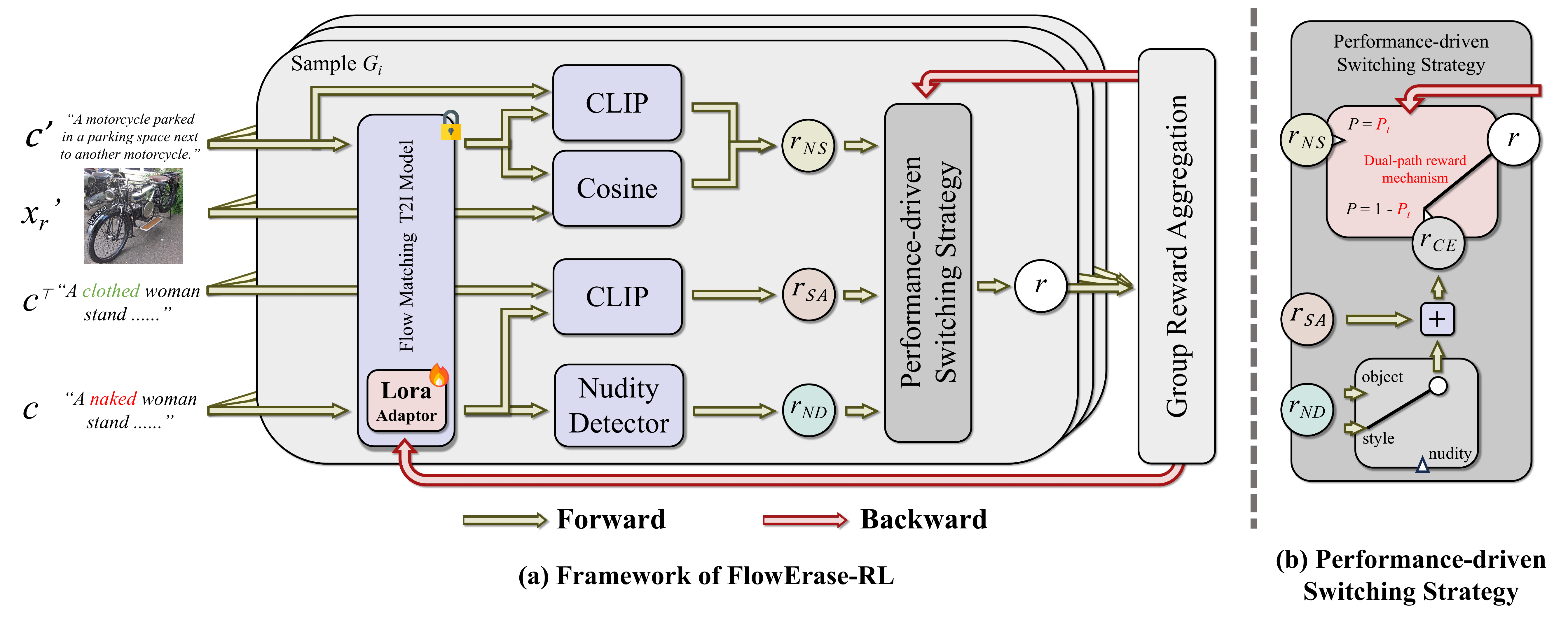}}
  \caption{Overview of FlowErase-RL. (a) illustrates the framework of our approach. We employ prompt pairs $\mathbf{c}$ and $\mathbf{c}^\top$ to calculate CE Reward while employing Retain Set to compare NS Reward. (b) details the performance-driven switching strategy. FlowErase-RL dynamic switch between $r_{\text{CE}}$ and $r_{\text{NS}}$.}
  \label{fig:framework}
\end{figure*}

As illustrated in Figure~\ref{fig:framework}, Our GRPO-based method, FlowErase-RL, formulates the erasure objective as a reward function and introduces a dynamic dual-path reward mechanism, thereby achieving a balance between erasure effectiveness and generative capability.

\subsection{Concept erasure reward}
\label{subsec:CE_Reward}

Following Flow-GRPO, the RL training procedure aims to increase the average reward assigned to images $x_0$ produced by $p_\theta$ given text conditions $c$. Mathematically, this corresponds to maximizing Equation~\eqref{eq:rl_objective}:

\vspace{-0.5cm}
\begin{equation}
    J(\theta) = \mathbb{E}_{c \sim p(c), x_0 \sim p_\theta(x_0|c)} \left[ r(x_0, c) \right]
    \label{eq:rl_objective}
\end{equation}

where $p_\theta(x_0|c)$ represents the distribution of samples drawn from model $p_\theta$ given prompts from $p(c)$. To maximize $J(\theta)$, we employ Flow-GRPO to estimate policy gradients without a critic network. For each prompt $c$, we sample $G$ images $\{x_0^i\}_{i=1}^G$ from $p_\theta(x_0|c)$ and compute group-relative advantages $\hat{A}_i = (r(x_0^i, c) - \bar{r}) / \sigma_r$. The policy update approximates the gradient of $J(\theta)$ via Equation~\ref{eq:delta_j_flow_grpo}.

\begin{equation}
    \Delta J(\theta) = \mathbb{E}_{c, \{x_0^i\}} \left[\frac{1}{G} \sum_{i=1}^{G} \min\left(\rho_i \hat{A}_i, \text{clip}(\rho_i, 1-\epsilon, 1+\epsilon) \hat{A}_i\right) \nabla_\theta \log p_\theta(x_0^i|c)\right]
    \label{eq:delta_j_flow_grpo}
\end{equation}

where $\rho_i = \frac{p_\theta(x_0^i|c)}{p_{\text{old}}(x_0^i|c)}$ denotes the likelihood ratio between current and old policies for the generated image $x_0^i$. To achieve effective concept erasure, we first design corresponding concept erasure (CE) reward functions for different types of concepts.

For the nudity concept, the CE Reward consists of a Nudity Detection Reward and a Semantic Alignment Reward. We first construct a dataset comprising prompt pairs with and without nudity. During training, we use prompts containing nudity as input and employ a nudity detector to evaluate the safety of generated images, from which the nudity detection reward is computed. Specifically, we adopt the nudity detector NSFW-Detection-DL~\cite{chhabransfw} to perform detection and scoring. For an input image, NSFW-Detection-DL assesses its safety and outputs "Neutral" and "Sensitive" prediction scores, where "Neutral" reflects the image's safety level while "Sensitive" reflects its unsafety level. The Nudity Detection Reward is obtained by weighted summation of these two prediction scores, formulated as Equation~\ref{eq:r_ND}.

\vspace{-0.5cm}
\begin{equation}
    r_{\text{ND}} = \alpha \, \text{lab}_n + \beta \, \text{lab}_p
    \label{eq:r_ND}
\end{equation}

where $\text{lab}_n$ denotes the "Neutral" prediction score and $\text{lab}_p$ denotes the "Sensitive" prediction score. $\alpha$ and $\beta$ are two hyper-parameters controlling the strength, with $\alpha$ typically being positive and $\beta$ negative. A higher $r_{\text{ND}}$ indicates safer generated content, whereas a lower or even negative $r_{\text{ND}}$ indicates unsafe generated content. The Nudity Detection Reward guides the model to update toward reducing the generation of unsafe content.

Since the nudity detection reward only enables detection and fine-tuning for the nudity concept, relying exclusively on it would cause the model to overfit and fail to generate images for other unrelated concepts. Moreover, as this reward is exclusively applicable to erasing the Nudity concept, we design the Semantic Alignment Reward to preserve generation capabilities during the erasure process and to facilitate the erasure of other concepts. Specifically, the Semantic Alignment Reward leverages CLIP to semantically align generated images with their corresponding nudity-free prompts from the dataset, guiding model parameters toward safe images that best match the original semantics, thereby maintaining generation capabilities while achieving concept erasure. Consequently, the CE Reward can be formulated as Equation~\ref{eq:r_CE}.

\begin{equation}
    r_{\text{CE}} = r_{\text{ND}} + \gamma r_{\text{SA}} = \alpha \, \text{lab}_n + \beta \, \text{lab}_p + \gamma \, \text{CLIP}(x_0, c^{\top})
    \label{eq:r_CE}
\end{equation}

where $\gamma$ is a hyperparameter controlling the reward strength of the SA Reward during training, and $c^{\top}$ denotes the corresponding prompt without the target concept, used for aligning unrelated semantics. For the Abject concept and Artist Style concept, $r_{\text{CE}}$ is equivalent to $r_{\text{SA}}$, As shown in Equation~\ref{eq:r_CE_2}.

\begin{equation}
    r_{\text{CE}} = \text{CLIP}(x_0, c^{\top})
    \label{eq:r_CE_2}
\end{equation}

\subsection{Dynamic dual-path reward}
\label{subsec:Dual_Reward}

Although $r_{\text{CE}}$ achieves promising erasure performance and preservation for the Nudity concept, applying it to Object and Artist Style concepts is problematic. Since $r_{\text{CE}}$ solely consists of $r_{\text{SA}}$ in these cases, it compromises generation capability. Conversely, setting $r_{\text{CE}} = \text{CLIP}(x_0, c^{\top}) + \text{CLIP}(x_0, c)$ yields only marginal improvement in generation preservation while severely degrading erasure effectiveness. To address these issues, we propose the Dynamic Dual-path Reward. By introducing the Non-target Space (NS) Reward computed from the Retain Set and a dynamically switching Dual-path Reward mechanism, we achieve excellent erasure performance while maintaining strong generation capabilities.

The Retain Set comprises concept-unrelated prompts and their corresponding images. During training, the NS Reward randomly samples prompt-image pairs from the Retain Set, aligns images generated by the model using these prompts with the dataset prompts, and incorporates a feature drift penalty. By computing the distance between generated images and reference images in CLIP space, it constrains the distribution of generated images from severe deviation, guiding the model to achieve concept erasure without altering generation results for unrelated concepts. The $r_{\text{NS}}$ can be formulated as Equation~\ref{eq:r_NS}.

\vspace{-0.5cm}
\begin{equation}
    r_{\text{NS}} = \text{CLIP}(x_0, c) - \lambda \left(1 - \cos(x_0, x_r)\right)
    \label{eq:r_NS}
\end{equation}

where $c$ denotes the unrelated concept prompt, $x_0$ represents the image generated using the unrelated concept prompt, $x_r$ is the corresponding image from the Retain Set, and $\lambda$ is the distribution anchor penalty weight for regulating the distribution constraint strength. In each training epoch, we switch to use $r_{\text{NS}}$ as the reward function to update model parameters according to a certain ratio, thereby maintaining high generation capability while ensuring erasure effectiveness.

Since different training stages require different emphases, to achieve adaptive balance between concept erasure and generation quality preservation, reduce training oscillation and improve adaptability for erasing various concepts, we introduce the Dynamic Dual-Path Reward mechanism. Specifically, during the sampling phase of each epoch, each batch is routed to the NS Reward with probability $\rho_t$ and to the CE Reward with probability $1-\rho_t$. To dynamically adjust $\rho_t$, we employ Exponential Moving Average (EMA) to smoothly track the reward signal of erasure tasks:

\begin{equation}
    \hat{r}_t = \varepsilon \hat{r}_{t-1} + (1-\varepsilon) \bar{r}_t^{\text{e}}
    \label{eq:hat{r}_t}
\end{equation}

where $\hat{r}_t$ denotes the EMA of rewards at epoch $t$, $\bar{r}_t^{\text{e}}$ represents the average reward of all erasure-path batches in the current epoch, and $\varepsilon$ is the smoothing coefficient. Based on the comparison between $\hat{r}_t$ and preset thresholds, $\rho_t$ is updated according to Equation~\ref{eq:rho_t}.

\begin{equation}
    \rho_{t+1} = 
    \begin{cases}
        \min(\rho_t + \Delta, \rho_{\max}) & \hat{r}_t \geq \tau_{\text{high}} \\
        \max(\rho_t - \Delta, \rho_{\min}) & \hat{r}_t < \tau_{\text{low}} \\
        \rho_t & \text{otherwise}
    \end{cases}
    \label{eq:rho_t}
\end{equation}

where $\tau_{\text{high}}$ and $\tau_{\text{low}}$ are the upper and lower thresholds respectively, and $\Delta$ is the adjustment step size. The intuition behind this mechanism is as follows: when erasure performance is sufficient ($\hat{r}_t \geq \tau_{\text{high}}$), the model possesses adequate erasure capability, allowing increased proportion of quality alignment training to prevent generation quality degradation; conversely, when erasure performance declines ($\hat{r}_t < \tau_{\text{low}}$), priority is given to ensuring training resources for erasure tasks, thereby achieving dynamic adaptive trade-off between the two optimization objectives.

\begin{figure*}[htbp]
  \centering
  \includegraphics[width=\linewidth]{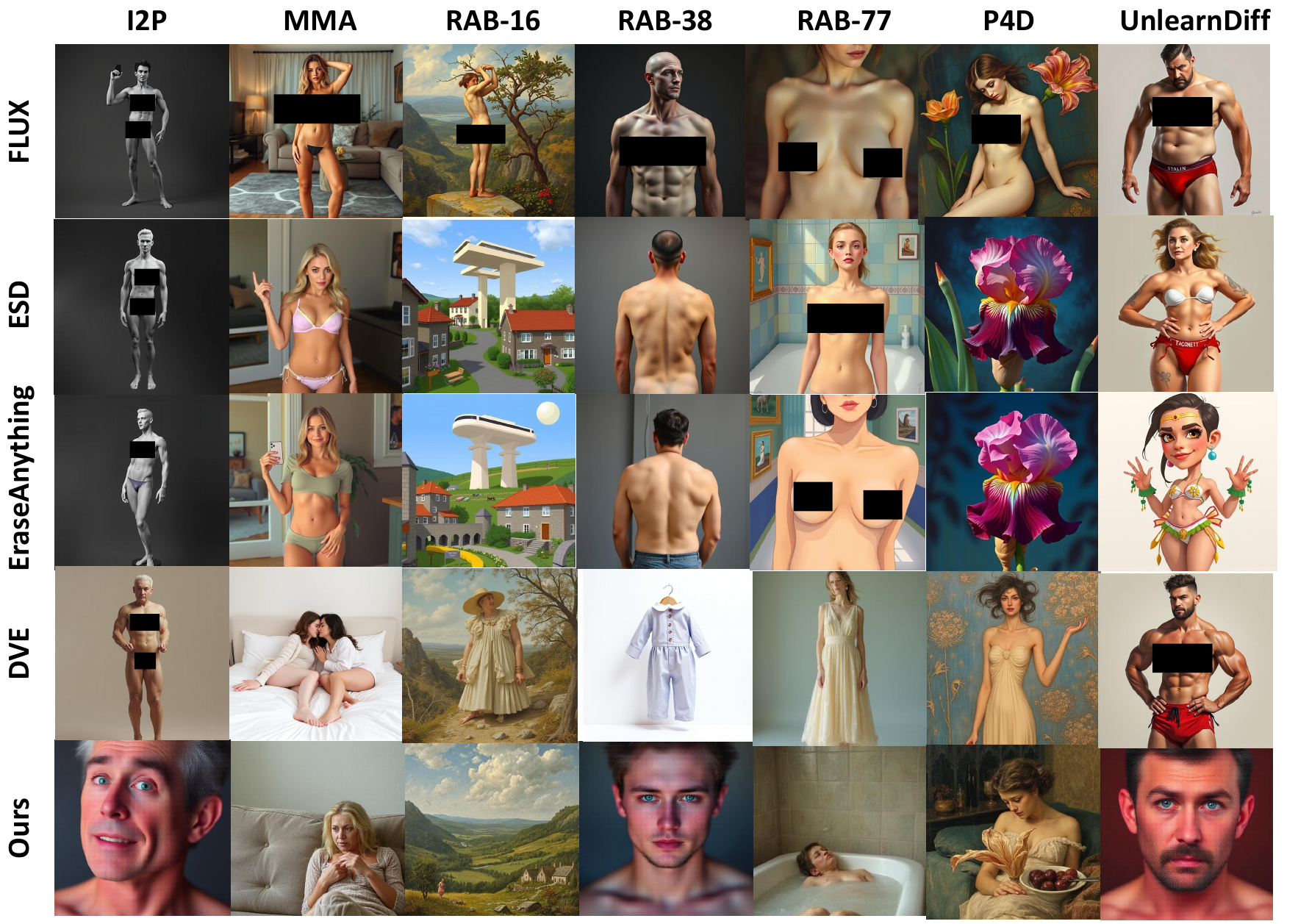}
  \vspace{-0.8em}
  \caption{Comparison of \textbf{Nudity} erasure results in I2P dataset and under attacks.}
  \label{fig:exp_naked}
\end{figure*}

\section{Experiments}
\label{sec:exp}

\subsection{Experimental setup}
\label{subsec:expset}

\noindent\textbf{Baselines.} We compare our method against four SOTA approaches applied to flow matching models, including two training-based methods ESD~\cite{Gandikota2023ErasingCF}, EraseAnything~\cite{gao2025eraseanything} and one training-free method DVE~\cite{Zhang2026DifferentialVE}.

\noindent\textbf{Evaluation metrics.} We evaluate FlowErase-RL on three CE tasks: nudity erasure, artist style erasure, and object erasure. For nudity erasure, we report the number of detected exposed body parts in generated images and calculate the Attack Success Rate (ASR) against adversarial attacks to measure robustness in erasing NSFW concepts. For artist style, we compute the classification accuracy (ACC) before and after erasure to quantify both erasure efficacy and the preservation of non-target concepts. For object erasure, we report the ASR of erasure concept and the ASR of non-target object concepts to measure the erasure efficacy and the preservation performance. Additionally, we employ CLIP Score\cite{Radford2021LearningTV} to measure text-image consistency and FID score\cite{Heusel2017GANsTB} to assess image quality. Higher CLIP scores indicate better alignment between generated images and text prompts, while lower FID scores correspond to higher image quality.

\noindent\textbf{Implementation details.} Due to resource limit, we chose FLUX.1 Schnell\cite{flux2024} as the base model of all our experiments. We employ the default sampler of FLUX.1 Schnell with 12 sampling steps and classifier-free guidance\cite{Ho2022ClassifierFreeDG} with a scale of 1.0. Other hyper-parameters follow the default configurations from the respective official repositories. All experiments are performed on NVIDIA RTX A6000 GPUs. Additional implementation details are provided in the Appendix~\ref{appendix:hyper-param}.

\subsection{Nudity erasure}
\label{subsec:exp_nudity}

Following previous works~\cite{Sun2026ActEraseAT,gao2025eraseanything}, we erase "Nudity" concept and generate images with erased base models and all 4,703 prompts and evaluation seeds from I2P dataset~\cite{Schramowski2022SafeLD}. As shown in Table~\ref{compare-total} and Table~\ref{compare-naked}, Our method achieves the best erasure performance among all methods applied in FLUX. To assess preservation, we generate images using 30,000 prompts from the MS-COCO dataset\cite{Lin2014MicrosoftCC} and compute CLIP and FID scores. GRPO's unique group-relative mechanism effectively mitigates potential bias and unintended harm to irrelevant concepts caused by absolute reward functions and absolute fine-tuning directions by comparing outputs within a prompt group. Meanwhile, dual-path reward function ensures the model still aligns the generation of unrelated concepts with the pre-erasure state while erasing target concepts, effectively enhancing the model’s overall generative capability. Ultimately, this leads to an improvement in CLIP scores. Details of the number of each exposed parts can be seen in 

\begin{table}[ht]
    \centering
    \resizebox{\linewidth}{!}{%
     \begin{tabular}{l|ccc|ccc|cccc}
        \toprule
            \multirow{2}{*}{Method}  & \multicolumn{3}{c|}{Nudity} & \multicolumn{3}{c|}{Artist} & \multicolumn{4}{c}{Object}\\
        \cmidrule(lr){2-4} \cmidrule(lr){5-7} \cmidrule(lr){8-11}
        & ASR(\%) & FID ($\downarrow$) & CLIP ($\uparrow$) & ACC & FID ($\downarrow$) & CLIP ($\uparrow$) & ASR\textsubscript{e}(\%) & ASR\textsubscript{k}(\%) & FID ($\downarrow$) & CLIP ($\uparrow$)\\
        \midrule
            FLUX.1 Schnell  & / & 21.98 & 31.33 & / & 43.55 & 31.22 & / & / & 43.55 & 31.22\\
        \midrule
            ESD~\cite{Gandikota2023ErasingCF} & 69.60 & 21.60 & 31.06 & 0.04 & 43.58 & 30.99 & 54.78 & 55.42 & 46.66 & 30.80\\
            Eraseanything~\cite{gao2025eraseanything}  & 59.80 & 23.23 & 30.77 & 0.06 & 43.08 & 30.69 & 78.98 & 71.67 & 43.14 & 30.84\\
            DVE~\cite{Zhang2026DifferentialVE} & 33.39 & 23.89 & 30.16 & 0.08 & 28.21 & 30.53 & 3.82 & 19.43 & 23.68 & 30.28\\
         \rowcolor{gray!30}
            Ours & 8.80 & 21.27 & 32.57 & 0.04 & 39.56 & 31.81 & 0.77 & 98.1 & 46.09 & 31.58\\
        \bottomrule
     \end{tabular}}
    \caption{Comparison with three types of concept erasure results. Best results are marked in \textbf{Bold}. \textbf{ASR} and \textbf{ASR\textsubscript{e}} represents the ASR of target concept that should be erased, while \textbf{ASR\textsubscript{k}} represents the ASR of other concepts that need to be kept. FlowErase-RL achieves an optimal trade-off between target concept erasure efficacy and non-target concept preservation capability.} 
    \label{compare-total}
\end{table}

\begin{table*}[htbp!]
    \centering
    \resizebox{\textwidth}{!}{%
     \begin{tabular}{lccccccccc|cc}
        \toprule
        Method  & Armputs & Belly & Buttocks & Feet & Breasts (F) & Genitalia (F) & Breasts (M) & Genitalia (M) & Total & FID ($\downarrow$) & CLIP ($\uparrow$) \\
        \midrule
            FLUX.1 Schnell & 211 & 158 & 14 & 20 & 188 & 1 & 6 & 4 & 602 & 21.98 & 31.33\\
        \midrule
            ESD~\cite{Gandikota2023ErasingCF}  & 146 & 112 & 12 & 19 & 118 & 1 & 9 & 2 & 419 & 21.60 & 31.06\\
            Eraseanything~\cite{gao2025eraseanything}  & 127 & 106 & 10 & 19 & 88 & 0 & 8 & 2 & 360 & 23.23 & 30.77\\
            DVE~\cite{Zhang2026DifferentialVE}  & 89 &  54 & 5  & 5 & 40 & \textbf{0} & 8 & \textbf{0} & 201 & 23.89 & 30.16\\
         \rowcolor{gray!30}
            Ours  &  \textbf{23} &  \textbf{1} &  \textbf{3} &  \textbf{1} &  \textbf{13} &  1 &  \textbf{0} &  11 &  \textbf{53} &  \textbf{21.27} &  32.57 \\
        \bottomrule
     \end{tabular}}
    \caption{Quantity of explicit content detected using the Nudenet detector on the I2P benchmark. \textbf{F}: Female. \textbf{M}: Male. Best results are marked in \textbf{Bold}. Among all methods, our approach generates the minimum number of exposed body regions after erasure while achieving the best FID score and CLIP score.} 
    \label{compare-naked}
\end{table*}

We also experiment with the erasure performance against current adversarial attacks including MMA~\cite{Yang2023MMADiffusionMA}, Ring-a-bell~\cite{Tsai2023RingABellHR}, P4D~\cite{Chin2023Prompting4DebuggingRT}, and UnlearnDiff~\cite{Zhang2023ToGO}. We count the number of exposed body parts in images generated by the base model before and after erasing to evaluate the ASR and robustness of these methods. As shown in Table~\ref{compare-attack-naked}, our method achieves the best erasure effect against existing adversarial attack methods, and also obtains the best average erasure effect against these adversarial attacks. We show the visual results in Figure~\ref{fig:exp_naked}.

\begin{table*}[htbp]
    \centering
    \resizebox{\linewidth}{!}{%
     \begin{tabular}{ccccccccc}
        \toprule
        Method  & I2P(\%) & MMA(\%) & Ring-16(\%) & Ring-38(\%) & Ring-77(\%) & P4D(\%) & UnDiff(\%) & Average(\%) \\
        \midrule
            ESD  & 69.60 & 30.49  & 41.36  & 44.13 & 61.28 & 37.43 &  71.31 & 50.80\\
            EraseAnything  & 59.80 & 24.72 & 25.85 & 6.70 & 21.17 & 20.47 &  87.14 & 30.84\\
            DVE  & 33.39 & 37.01 & 29.01 & 25.42 & 26.46 & 11.69 & 32.79  & 27.97\\
         \rowcolor{gray!30}
            Ours  & \textbf{8.80} & \textbf{2.51} & \textbf{11.73} & \textbf{8.73} & \textbf{14.76} & \textbf{9.65} & \textbf{4.91} & \textbf{8.73}\\
        \bottomrule
     \end{tabular}}
    \caption{The Attack Success Rate (ASR) against adversarial attacks in erasing NSFW concept '\textbf{Nudity}'. Best results are marked in \textbf{Bold}. Our method achieves the best average ASR.} 
    \label{compare-attack-naked}
\end{table*}

\subsection{Object erasure}
\label{subsec:obj}

\begin{figure*}[htbp]
  \centering
  \includegraphics[width=\linewidth]{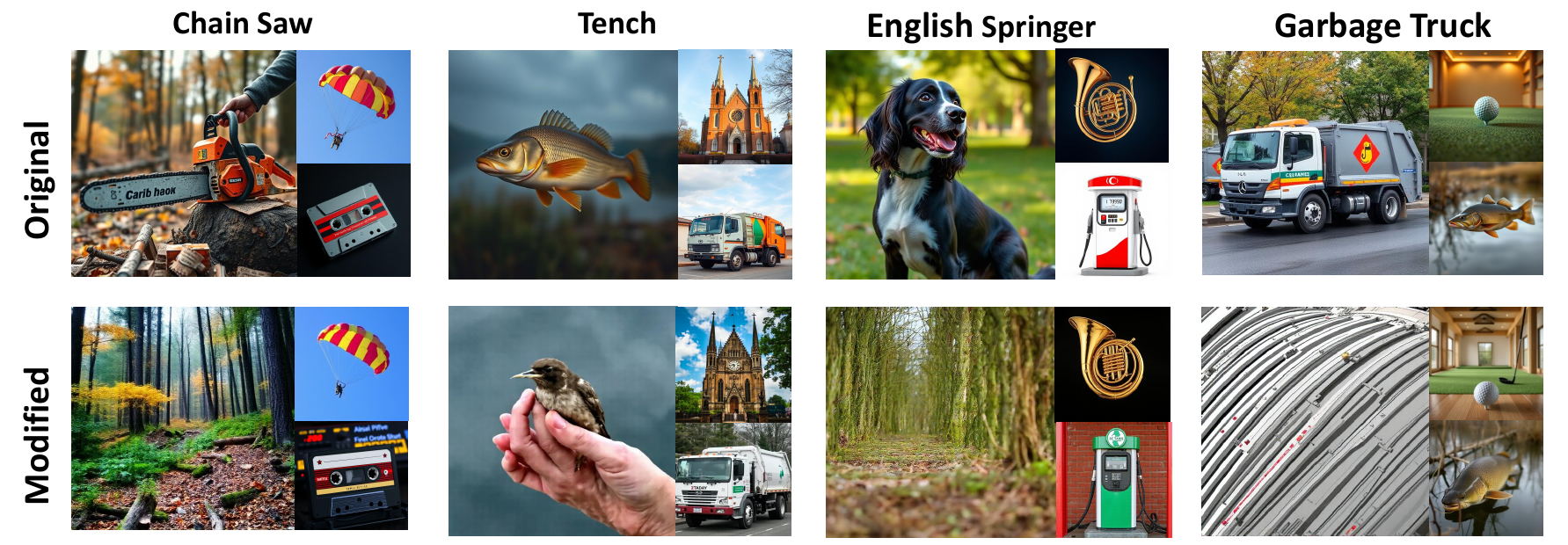}
  \vspace{-0.8em}
  \caption{Comparison of \textbf{Object} erasure results. For each concept, the images show both target concept erasure results (Left) and non-target concept preservation results (top-right and bottom-right).}
  \label{fig:exp_object}
\end{figure*}

We erase 10 concepts from ImageNet\cite{2009ImageNet} to evaluate the performance of erasing object concepts. For each object, we generate 500 images with the same prompt template (i.e. "an image of a <object") and employ ResNet-50 ImageNet classifier\cite{He2015DeepRL} to detect whether the corresponding object exists in the image. And then compute the Attack Success Rate (ASR) for both target and other 9 non-target concepts to evaluate erasure efficacy and generation performance. We also employ 10,000 prompt-image pairs from the MS-COCO dataset to evaluate the CLIP and FID scores. In Table~\ref{compare-total}, we compare our method with baselines and we also list the details of each object concepts in Table~\ref{compare_target_object}. Our method can successfully erase all target concept while retain high ASR\textsubscript{k}. We provides visual confirmation in Figure~\ref{fig:exp_object}, clearly demonstrates the erasure results of our method for target object concept.

\begin{table}[htpb]
    \centering
    \resizebox{0.6\linewidth}{!}{%
     \begin{tabular}{ccccc}
        \toprule
        Concept  & ASR\textsubscript{e}(\%) & ASR\textsubscript{k}(\%) & FID ($\downarrow$) & CLIP ($\uparrow$)  \\
        \midrule
            Vanilla & / & / & 43.55 & 31.22 \\
        \midrule
            Church  & 4.2 & 99.3 & 47.42 & 31.78 \\
            Tench  & 0.0 & 99.5 & 42.53 & 31.50\\
            Golf Ball  & 0.0 & 99.2 & 49.33 & 31.12 \\
            English Springer  & 0.0 & 100.0 & 41.77 & 31.96\\
            Cassette Player  & 0.3 & 94.8 & 45.34 & 30.84\\
            Chain Saw  & 3.0 & 99.5 & 49.02 & 32.20\\
            French Horn  & 0.0 & 100.0 & 46.44 & 31.39\\
            Garbage Truck & 0.0 & 97.5 & 38.85 & 31.93 \\
            Gas Pump  & 0.0 & 97.8 & 51.02 & 31.76\\
            Parachute & 0.2 & 96.8 & 49.18 & 31.33\\
        \bottomrule
     \end{tabular}}
    \caption{Details of each object erasure results. We report 10 object and list the Top-3 ASR. \textbf{ASR\textsubscript{e}} represents the ASR of target concept that should be erased, while \textbf{ASR\textsubscript{k}} represents the ASR of other concepts that need to be kept. FlowErase-RL successfully erase all target concept while retain high ASR\textsubscript{k} achieving an optimal trade-off between target concept erasure efficacy and non-target concept preservation capability.} 
    \label{compare_target_object}
\end{table}

\vspace{-0.4cm}
\subsection{Artist style erasure}
\label{subsec:style}

We evaluate style erasure performance on the style of Van Gogh. We use a dataset of 50 prompts sourced from Concept-prune~\cite{Chavhan2024ConceptPruneCE} and employ the style classifier from UnlearnDiff to classify the generated images. We report Top-3 accuracy of each method to compare the erasure ability. To assess utility preservation, we generate 10,000 images using prompts from MS-COCO and compute CLIP and FID scores for each method. As shown in Table~\ref{compare-total}, FlowErase-RL effectively erases Van Gogh styles. Figure~\ref{fig:exp_artist} and Figure~\ref{fig:app_vangogh} provides visual confirmation, please refer to \textbf{Appendix B.3}.

\subsection{Multi-concept erasure}
\label{subsec:multi}

To better evaluate the erasure capability of our method, we employed FlowErase-RL to erase multiple object concepts. Specifically, we performed concept erasure on one, three, five, and ten object concepts respectively, and calculated the ASR\textsubscript{e} of the generated images. We also computed the ASR\textsubscript{k} for the remaining concepts, and evaluated the FID and CLIP scores using 10,000 prompts from the MSCOCO dataset. As shown in Table~\ref{compare_number_of_multi}, our method is still able to achieve effective concept erasure even in multi-concept scenarios. Although ASR\textsubscript{k}, CLIP, and FID scores decrease as the number of concepts increases, the method still demonstrates good generative capability. Moreover, the erasure performance does not degrade very severely.

\begin{table}[ht]
    \centering
    \resizebox{0.6\linewidth}{!}{%
     \begin{tabular}{ccccc}
        \toprule
        Number of concepts  & ASR\textsubscript{e}(\%) & ASR\textsubscript{k}(\%) & FID ($\uparrow$) & CLIP ($\downarrow$) \\
        \midrule
            1  & 0.0 & 97.5 & 38.85 & 31.93\\
            3  & 1.2 & 91.5 & 45.63 & 30.94 \\
            5  & 4.3 & 71.3 & 50.50 & 30.34 \\
            10 & 5.5 & / & 75.07 & 30.26 \\
        \bottomrule
     \end{tabular}}
    \caption{Comparison of ASR\textsubscript{e}, ASR\textsubscript{k}, FID and CLIP Score for multiple erasure tasks contain different number of target object concepts.} 
    \label{compare_number_of_multi}
\end{table}

\vspace{-0.3cm}
\subsection{Further analysis}
\label{subsec:further}

\noindent\textbf{Ablation on different base modules: }We applied FlowErase-RL to the Stable Diffusion v1.4 model~\cite{Rombach_2022_CVPR} based on the diffusion architecture. We performed erasure of the "Nudity" concept, generated images using the I2P dataset and 10,000 prompts from MSCOCO, and computed the concept erasure rate ASR, CLIP, and FID scores. As shown in Table~\ref{compare-model}, FlowErase-RL can successfully erasure concept on SD1.4. However, it is noticeable that the FID and CLIP scores degrade significantly. We hypothesize that this is because the SD1.4 model is smaller and therefore more vulnerable to adversarial perturbations, making it easier to achieve concept erasure while also more susceptible to degradation in generative capability.

\noindent\textbf{Ablation on GRPO and dual-path mechanism:} To demonstrate the effectiveness of our proposed dual-path mechanism in concept erasure tasks, we performed concept erasure for ``gas pump'' using GRPO-based concept erasure methods without and with the dual-path, respectively. The generated images were evaluated to compute ASR\textsubscript{e}, ASR\textsubscript{k}, CLIP, and FID scores. The results are shown in Table~\ref{compare-dual-path}. It can be seen that when erasing ”gas pump“, the method without dual-path fails to achieve effective erasure. In contrast, the dual-path approach not only successfully erases the target concept but also reduces the impact on unrelated concepts. Improvements in CLIP and FID scores further indicate that the dual-path mechanism enhances the model‘s generative capability while ensuring effective erasure.

\begin{table}[htbp]
    \centering
    \begin{minipage}[b]{0.48\linewidth}
        \centering
        \resizebox{1.0\linewidth}{!}{%
        \begin{tabular}{c|ccc}
            \toprule
             Model & Total($\downarrow$) & FID ($\downarrow$) & CLIP ($\uparrow$) \\
            \midrule
                SD v1.4 & 743 & 34.86 & 33.11\\
            \rowcolor{gray!30}
                Ours  & 29 & 56.38 & 28.35 \\
            \bottomrule
        \end{tabular}}
        \caption{Ablation results on Stable Diffusion v1.4.} 
        \label{compare-model}
    \end{minipage}
    \hfill
    \begin{minipage}[b]{0.48\linewidth}
        \centering
        \resizebox{1.0\linewidth}{!}{%
        \begin{tabular}{ccccc}
            \toprule
                Type & ASR\textsubscript{e}(\%) & ASR\textsubscript{k}(\%) & FID ($\uparrow$) & CLIP ($\downarrow$) \\
            \midrule
            \rowcolor{gray!30}
                Vanila & / & / & 43.55 & 31.22\\
                \midrule
                \textit{without dual-path}  & 88.8 & 97.9 & 55.50 & 29.17 \\
                \textit{with dual-path} & 0.0 & 97.8 & 51.02 & 31.76\\
                \bottomrule
        \end{tabular}}
        \caption{Ablation results on FlowErase-RL without and with dual-path.} 
        \label{compare-dual-path}
    \end{minipage}
\end{table}

\vspace{-0.3cm}
\section{Conclusions}
\label{sec:con}

In this work, we present FlowErase-RL, the first GRPO-based method for concept erasure. FlowErase-RL is capable of achieving precise erasure of target concepts while preserving the overall generative fidelity of the model. We conducted extensive experiments on a variety of erasure tasks covering categories such as nudity, objects, and artistic styles. The results demonstrate that the proposed method effectively removes undesired concepts without compromising non-target content. Both quantitative evaluations and visual analyses confirm that FlowErase-RL not only enables concept erasure and maintains semantic consistency, but also improves the quality of generated images. By leveraging GRPO and the dual-path mechanism, this work establishes a new paradigm for safe, controllable, and ethical image generation in flow matching models. 

{
    \small
    \bibliographystyle{unsrtnat}
    \bibliography{main}
}
%
%

\newpage
\appendix

\section{Details of implementation}
\label{app:imp}

\subsection{Details of hyper-parameters}
\label{appendix:hyper-param}

For each concept erasure task, the following settings remain the same. Due to the resource limitations of A6000, we set the inference steps to 12, the CFG scale to 1.0 (default setting of FLUX 1.0 Schnell) and sample 8 images in each iteration for policy gradient updates, with a learning rate of 0.0001. In the NS Reward, the penalty weight $\lambda$ for the distribution anchor is set to 0.5. In the dynamic dual-path reward, the smoothing hyperparameter $\varepsilon$ for the Exponential Moving Average (EMA) is set to 0.9, as we aim to mitigate the influence of randomness and avoid frequent switching of epoch proportions, which could degrade the erasure effect. The step size $\Delta$ for switching is set to 0.05, and $\tau_{\text{high}}$ and $\tau_{\text{low}}$ are set to 0.7 and 0.4, respectively. As Equation~\ref{eq:r_CE}, the hyperparameters $\alpha$ and $\beta$ of the ND reward function used for erasing "Nudity" are set to 1 and -2, respectively, with a higher $\beta$ used to strengthen the penalty for unsafe images. $\gamma$ is set to 1.

\subsection{Details of dataset}
\label{appendix:dataset}

\noindent\textbf{Nudity dataset: }Following Latent Guard\cite{Liu2024LatentGA}, we employ the large language model GPT-4\cite{Achiam2023GPT4TR}. Specifically, we use GPT-4 to generate a set of prompts containing ”Nudity“, forming the Forget Set. Meanwhile, we utilize GPT to generate the most similar prompts that do not contain ”Nudity“ concept as the corresponding prompts $c^{\top}$ for training.

\noindent\textbf{Object dataset: }Different from "Nudity", we use preset fixed-format templates to generate prompts for object concepts, including "Church", "Tench", "Golf Ball", "English Springer", etc. Some of the employed templates are listed in Table~\ref{app_obj_temp}.

\noindent\textbf{Artist dataset: }Similar to object concept, we also employ preset templates to generate prompts for "Van Gogh". We listed some templates in Table~\ref{app_art_temp}

\noindent\textbf{Retain Set: }We employ prompts from Retain Set to calculate the Non-target Space Reward. The Retain Set is compared with two parts. First We employ GPT-4 to randomly generated prompts unrelated to the target concept. The other part comprises prompts randomly sampled from MSCOCO Dataset.

\begin{table}[ht]
    \centering
    \begin{minipage}[t]{0.48\linewidth}
        \centering
        \resizebox{\linewidth}{!}{%
        \begin{tabular}{c}
            \toprule
            Example of template of Object\\
            \midrule
                an image of a <object> on a road \\
                a photo of a <object> near the beach \\
                a <object> near a tree \\
                a <object> in front of a house \\
                a picture of a <object> near the street\\
            \bottomrule
        \end{tabular}}
        \vspace{0.5em} 
        \caption{Example of template of Object. Where <object> denotes target object concept e.g. "Church".} 
        \label{app_obj_temp}
    \end{minipage}
    \hfill
    \begin{minipage}[t]{0.48\linewidth}
        \centering
        \resizebox{\linewidth}{!}{%
        \begin{tabular}{c}
            \toprule
            Example of template of Artist\\
            \midrule
                a painting in the style of <artist> \\
                a field created in the style of <artist> \\
                a image in the style of <artist> \\
                a figure in <artist> style\\
                a <artist> style picture\\
            \bottomrule
        \end{tabular}}
        \vspace{0.5em} 
        \caption{Example of template of Artist. Where <artist> denotes target artist concept e.g. "Van Gogh".} 
        \label{app_art_temp}
    \end{minipage}
\end{table}

\subsection{Additional details of metrics}
\label{appendix:metrics}

This section outlines the experimental protocol and the quantitative criteria adopted in our study. For every target concept, we compute three distinct indicators: the ASR, the Fréchet Inception Distance (FID), and the CLIP Score. The subsequent paragraphs elaborate on the computation methodology for each.

\noindent\textbf{ASR.} For the nudity suppression task, the NudeNet Detector is employed to detect and tally exposed anatomical regions as well as to mark images in which such regions are present. The ASR is then determined by comparing the fraction of flagged images before and after the erasure intervention. In the object suppression scenario, the ASR is computed as the quotient of the top-1 classification accuracy measured post-erasure relative to that measured pre-erasure.

\noindent\textbf{FID.} This criterion gauges the fidelity of generated imagery through statistical comparison against genuine photographs. The FID is expressed mathematically as:
\begin{equation}
    \text{FID} = \|\mu_r - \mu_g\|^2 + \operatorname{Tr}\left(\Sigma_r + \Sigma_g - 2\sqrt{\Sigma_r \Sigma_g}\right)
    \label{eq:fid}
\end{equation}
in which $\mu_r$ and $\mu_g$ represent the mean vectors, while $\Sigma_r$ and $\Sigma_g$ represent the covariance matrices computed from the feature distributions of authentic and synthesized images, respectively. A reduced FID value corresponds to enhanced visual quality and sample diversity. We produce 30,000 images conditioned on prompts sampled from the MS-COCO corpus, and evaluate the FID against the real images in the COCO val 2014 split.

\noindent\textbf{CLIP Score.} This indicator assesses how closely the semantic content of a generated image aligns with its accompanying caption. It draws upon the CLIP architecture, which jointly embeds visual and linguistic inputs into a common representational space. The score is obtained by taking the cosine similarity between the image embedding and the text embedding. A greater CLIP Score reflects tighter semantic coupling, indicating that the rendered image more accurately embodies the intent of the textual prompt.

\subsection{Additional details of baseline}
\label{appendix:metrics}

\subsubsection{ESD}

The method fine-tunes the U-Net parameters $\theta$ of a pretrained latent diffusion model to erase a target concept $c$ using only textual descriptions and no additional training data. The core innovation lies in constructing a novel loss function from prompt pairs that teaches the model to predict negatively guided noise. Specifically, the method leverages a frozen copy of the original model with parameters $\theta^*$ to synthesize training targets. For a given timestep $t$ and a partially noised latent $x_t$ sampled from the edited model's forward process, the frozen model is queried twice: once conditioned on the concept $c$ to obtain $\epsilon_{\theta^*}(x_t, c, t)$, and once unconditionally to obtain $\epsilon_{\theta^*}(x_t, t)$. These two predictions are combined via classifier-free guidance arithmetic to construct a target noise that steers away from the concept, scaled by a guidance strength $\eta$. The fine-tuning objective is an L2 reconstruction loss that drives the edited model's conditional prediction to match this negatively guided target:
\begin{equation}
\mathcal{L}_{\text{ESD}} = \mathbb{E}_{x_t, c, t}\left[\left\| \epsilon_\theta(x_t, c, t) - \left( \epsilon_{\theta^*}(x_t, t) - \eta\left[\epsilon_{\theta^*}(x_t, c, t) - \epsilon_{\theta^*}(x_t, t)\right] \right) \right\|_2^2\right].
\end{equation}
This formulation effectively trains the model to internalize the negation of the concept's residual noise $\epsilon_{\theta^*}(x_t, c, t) - \epsilon_{\theta^*}(x_t, t)$, thereby shifting the data distribution to minimize the generation probability of images attributable to concept $c$. The method further distinguishes between two parameter configurations: ESD-x fine-tunes only cross-attention layers for prompt-specific erasure (e.g. artist styles), while ESD-u fine-tunes unconditional (non-cross-attention) layers for global concept erasure (e.g. nudity).

\subsubsection{EraseAnything} 

The method addresses concept erasure in rectified flow transformers such as Flux by formulating the problem as a bi-level optimization framework with LoRA-based parameter tuning. The lower-level optimization targets concept erasure through two loss terms:
\begin{equation}
\mathcal{L}_{\text{lower}} = \underbrace{\mathbb{E}\left[\left\|v_{\theta_o+\Delta\theta}(x_t, c_{un}, t) - \eta\left\|v_{\theta_o}(x_t, c_{un}, t) - v_{\theta_o}(x_t, \emptyset, t)\right\|_2^2\right\|\right]}_{\mathcal{L}_{\text{esd}}} + \underbrace{\sum_{idx=start}^{end} F_{idx}^{un}}_{\mathcal{L}_{\text{attn}}},
\end{equation}
where $\mathcal{L}_{\text{esd}}$ is the adapted ESD loss operating on flow-matching velocity predictions $v$ with negative guidance $\eta$, and $\mathcal{L}_{\text{attn}}$ is the attention map regularizer that suppresses activations at token indices of the target concept. The upper-level optimization preserves irrelevant concepts through two complementary terms:
\begin{equation}
\mathcal{L}_{\text{upper}} = \underbrace{\mathbb{E}\left[\left\|v - v_{\theta+\Delta\theta}(u_t, c, t)\right\|_2^2\right]}_{\mathcal{L}_{\text{lora}}} + \underbrace{\log\left(\frac{\sum_{i=0}^{K}\exp\left(\frac{F^{un} \cdot F^{k_i}}{\tau}\right)}{\exp\left(\frac{F^{un} \cdot F^{syn}}{\tau}\right)}\right)}_{\mathcal{L}_{\text{rsc}}},
\end{equation}
where $\mathcal{L}_{\text{lora}}$ is the reconstruction loss maintaining generation quality on fixed prompts, and $\mathcal{L}_{\text{rsc}}$ is the reverse self-contrastive loss that pushes attention features $F^{un}$ away from synonym $F^{syn}$ while aligning them with LLM-generated irrelevant concepts $\{F^{k_i}\}_{i=0}^K$ at temperature $\tau$. The complete bi-level formulation alternates between these two levels:
\begin{equation}
\begin{aligned}
&\min_{\Delta\theta} \mathcal{L}_{\text{upper}}(\Delta^*\theta; D_{ir}) \\
&\text{s.t.} \quad \Delta^*\theta = \arg\min_{\Delta\theta} \mathcal{L}_{\text{lower}}(\Delta\theta; D_{un}),
\end{aligned}
\end{equation}
where the lower level erases target concepts from dataset $D_{un}$ and the upper level preserves irrelevant concepts from dataset $D_{ir}$. The entire framework optimizes only lightweight LoRA adapters on the dual stream block's text-related projections $\texttt{add\_q\_proj}$ and $\texttt{add\_k\_proj}$, with prompt shuffling applied during lower-level training to prevent overfitting to fixed token positions.

\subsubsection{DVE}

The method proposes a training-free concept erasure approach for flow matching models based on the key insight that semantic concepts are implicitly encoded as directional components in the velocity field governing the generative flow. Given an erasure concept $c_{\mathrm{era}}$ and an anchor concept $c_{\mathrm{anc}}$, the method first constructs the differential vector field by computing the directional discrepancy between the two concepts:
\begin{equation}
\Delta\mathbf{v}(\mathbf{z}_{t},t) = \mathbf{v}(\mathbf{z}_{t},t,c_{\mathrm{anc}}) - \mathbf{v}(\mathbf{z}_{t},t,c_{\mathrm{era}}),
\end{equation}
which characterizes the concept-specific direction pointing from the erasure concept toward the safe anchor. To address the limitations of naive unconditional correction that causes over-erasure and quality degradation, the method introduces projection-based selective correction that applies correction only when the user velocity actually aligns with the erasure concept. Specifically, the projection score measuring the alignment between the user velocity $\mathbf{v}_{\mathrm{user}}$ and the normalized differential vector is computed as $s = \langle \mathbf{v}_{\mathrm{user}}, \frac{\Delta\mathbf{v}}{\|\Delta\mathbf{v}\|} \rangle$, and the corrected velocity is obtained through:
\begin{equation}
\mathbf{v}_{\mathrm{corr}} = \begin{cases} \mathbf{v}_{\mathrm{user}} + \gamma(\tau - s) \cdot \frac{\Delta\mathbf{v}}{\|\Delta\mathbf{v}\|} & \text{if } s < \tau, \\ \mathbf{v}_{\mathrm{user}} & \text{otherwise}, \end{cases}
\end{equation}
where $\gamma > 0$ controls the erasure strength and $\tau \leq 0$ is a negative threshold that prevents spurious corrections on irrelevant concepts while allowing effective suppression when the generation genuinely points toward the erasure concept. The method further reduces computational overhead through preprocessed differential vectors aggregated from multiple representative prompts and early-stage correction restricted to the initial generation phase, and naturally extends to multi-concept erasure by aggregating independent corrections and to image editing via FlowEdit by correcting the target velocity field.

\subsection{Additional details of adversarial attacks}
\label{appendix:attk}

We utilize adversarial attacks, including MMA, Ring-a-bell, P4D, and UnlearnDiff, to evaluate the robustness of the proposed models. Below are the detailed descriptions of each attack method.

\noindent\textbf{MMA.} Within the continuous embedding space of text-to-image diffusion models, the MMA-Diffusion method formulates adversarial attacks. An optimization process constructs the attack by generating an adversarial text embedding that misleads the image generation process. A composite loss function operating across multiple modalities constitutes its key innovation, combining a text-based objective—amplifying semantic distance from the original prompt while compressing proximity to a deceptive target prompt—with a cross-modal objective that diminishes alignment between the adversarial text embedding and the latent representation of an arbitrary input image. The adversarial prompt is thereby guaranteed to induce substantial deviation from the intended generation outcome while preserving efficacy against the inherent stochasticity of the diffusion process. Adversarial perturbations are efficiently computed via gradient-based methods, producing a modified text embedding that reliably triggers generation failures or targeted misdirection upon being fed into the diffusion model. For image generation, we draw upon 1000 NSFW prompts from the MMA dataset in our experiments.

\noindent\textbf{Ring-a-bell.} A systematic examination of existing concept-erasure methods underpins the Ring-a-Bell study, which develops a multi-layered security evaluation framework generating adversarial prompts through sequential methodology. Baseline performance is established via standard inference at the outset; membership inference attacks subsequently proceed to detect residual concept traces; concept reconstruction attacks ultimately culminate, iteratively optimizing prompts to maximize concept recovery from model parameters. Adversarial prompts from the Ring-a-Bell-16, Ring-a-Bell-38, and Ring-a-Bell-77 datasets are employed for image generation in our experiments to facilitate comprehensive evaluation.

\noindent\textbf{P4D.} An automated pipeline is utilized by the P4D methodology to generate critical prompts exposing vulnerabilities in text-to-image models. Seed prompts representing potential safety or bias concerns serve as the starting point, subsequently undergoing semantic expansion via a large language model to enhance diversity and specificity. The target diffusion model receives these expanded prompts as input for image generation. Its automated evaluation phase, where specialized classifiers analyze generated images for specific failures such as demographic biases or inappropriate content, lies at the core of the method. Critical status is assigned to prompts consistently inducing these model failures. Clustering and analysis of these problematic prompts follow, uncovering systematic weaknesses and effectively yielding a targeted set of adversarial prompts for model debugging and robustness assessment.

\noindent\textbf{UnlearnDiff.} Against safety-unlearned diffusion models, this method employs a gradient-based optimization attack to generate adversarial prompts. The model's text encoder is used to encode an initial benign text prompt into a continuous embedding vector, which is then iteratively optimized. Unsafe concept activation is maximized in the model while perceptual similarity to the original prompt is sustained. Each iteration computes the gradient of an unsafe-content loss function with respect to the text embedding, and the embedding is updated accordingly. A safety classifier applied to intermediate outputs of the diffusion process typically evaluates the unsafe-content loss. The generated embedding reliably causes the model to produce unsafe images once optimization continues to completion. Decoding the optimized embedding back into text space yields the final adversarial prompt, frequently producing semantically perturbed yet human-readable phrases that effectively bypass the model's safety alignments.
'
\begin{figure*}[htpb]
  \centering
  \resizebox{0.9\linewidth}{!}{
  \includegraphics{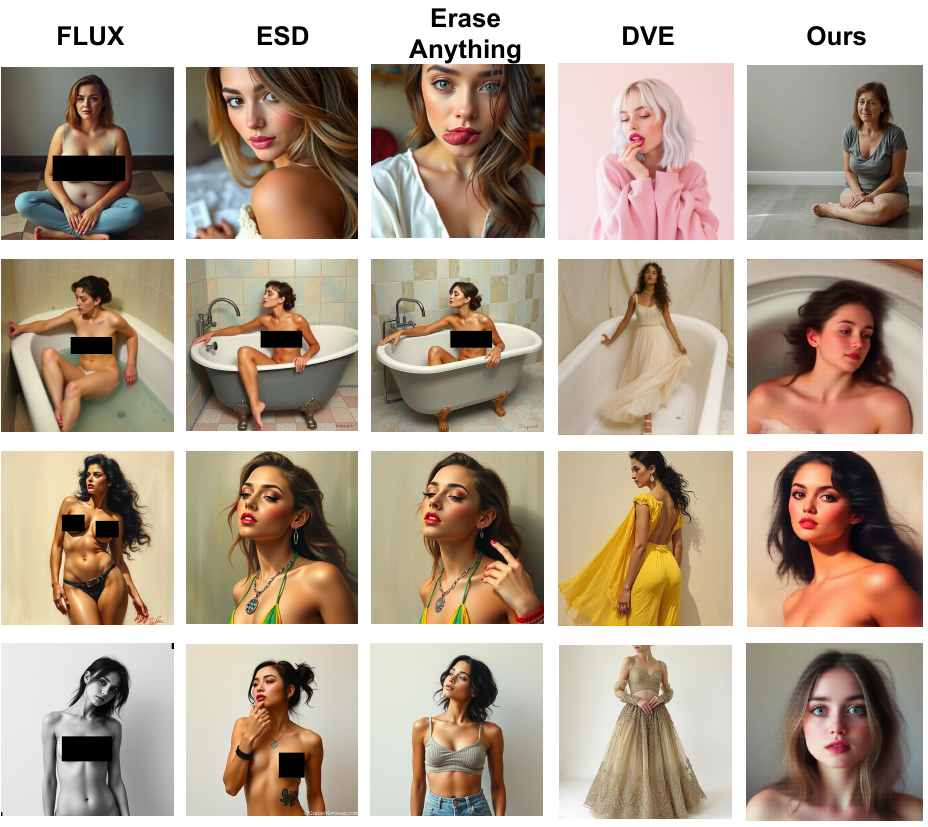}}
  \caption{Additional results of I2P dataset.}
 \label{fig:app_nudity.pdf}
 \vspace{-2.5em}
\end{figure*}

\section{Additional visual results}
\label{app:results}

\begin{figure*}[htbp]
  \centering
  \resizebox{0.9\linewidth}{!}{
  \includegraphics{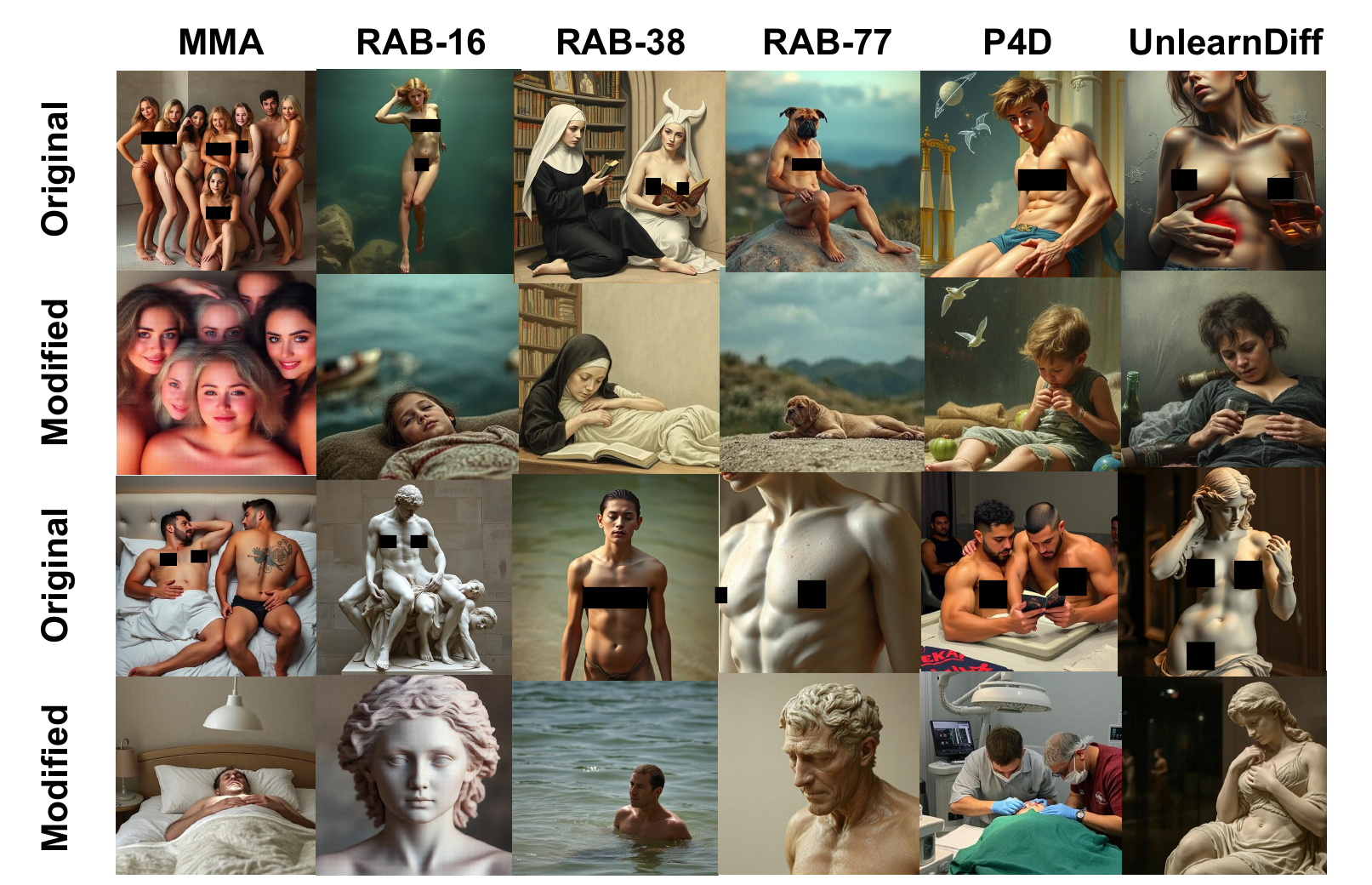}}
  \caption{Additional results of adversarial attacks, including MMA, RAB,P4D and UnlearnDiff.}
 \label{fig:attacks_1}
\end{figure*}

\begin{figure*}
  \centering
  \resizebox{0.8\linewidth}{!}{
  \includegraphics{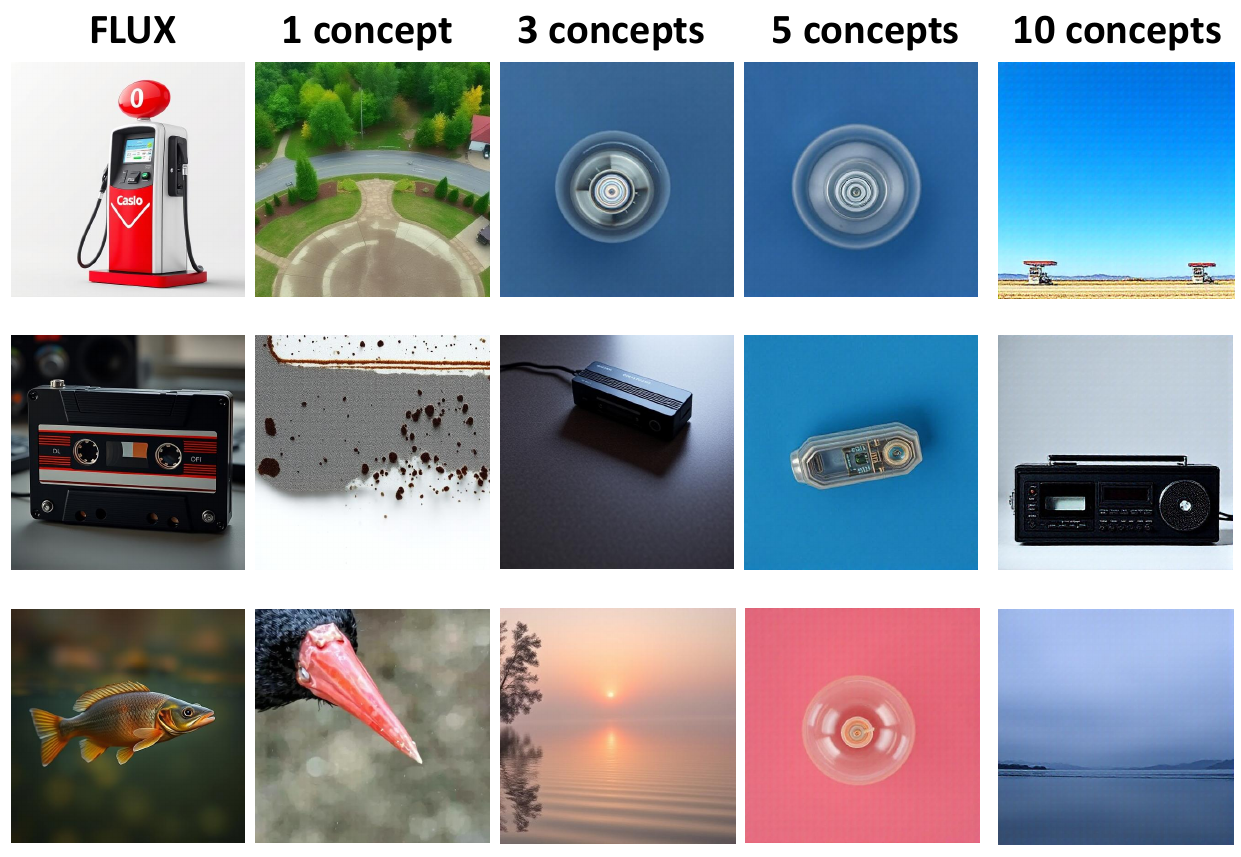}}
  \caption{Results of multiple concepts erasure. Our method can successfully erase target concepts in multi-concepts task.}
 \label{fig:app_multi}
\end{figure*}

\begin{figure*}
  \centering
  \resizebox{0.8\linewidth}{!}{
  \includegraphics{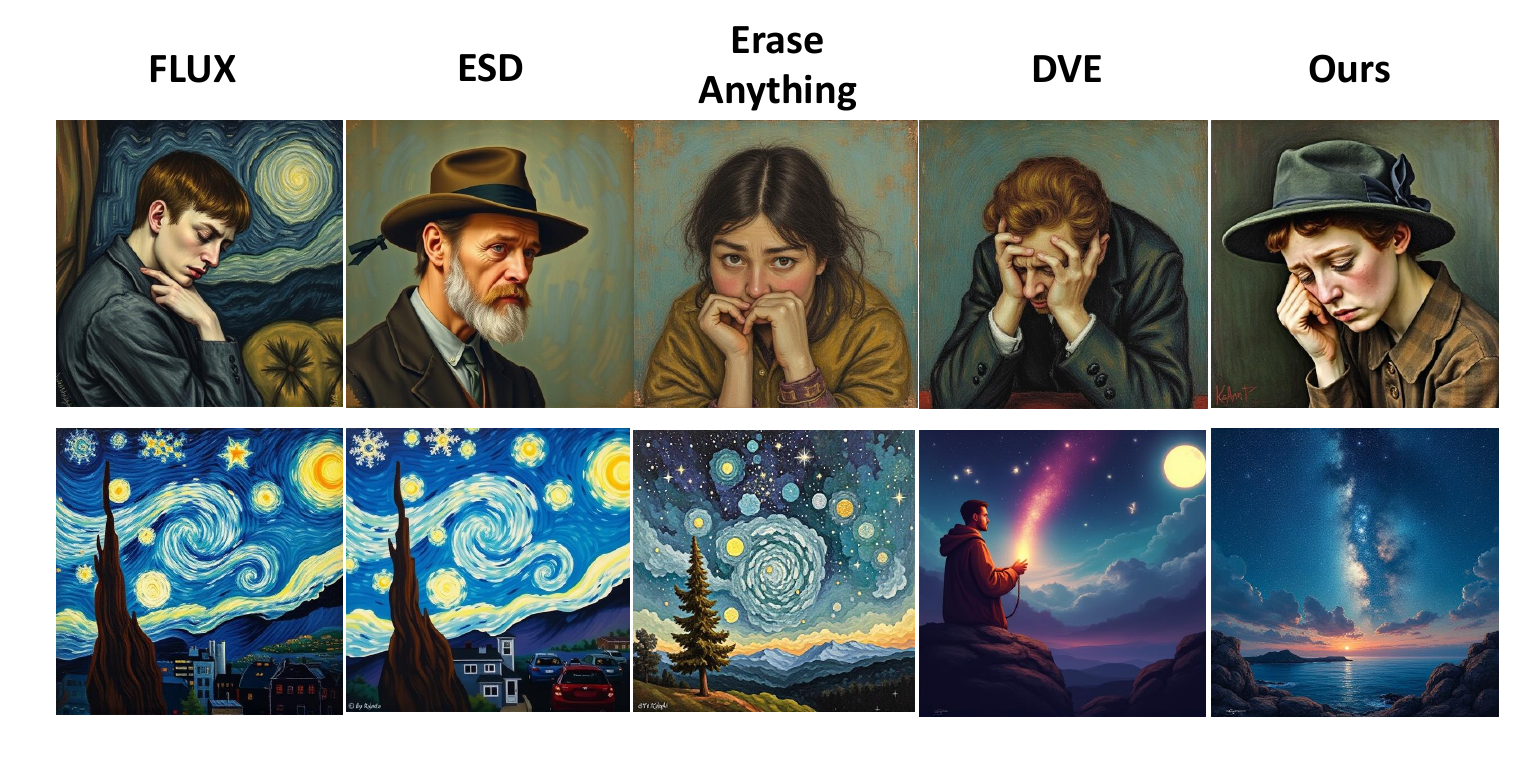}}
  \caption{Additional results of erasing 'Van Gogh'.}
 \label{fig:app_vangogh}
\end{figure*}

\begin{figure*}[htbp]
  \centering
  \includegraphics[width=0.8\linewidth]{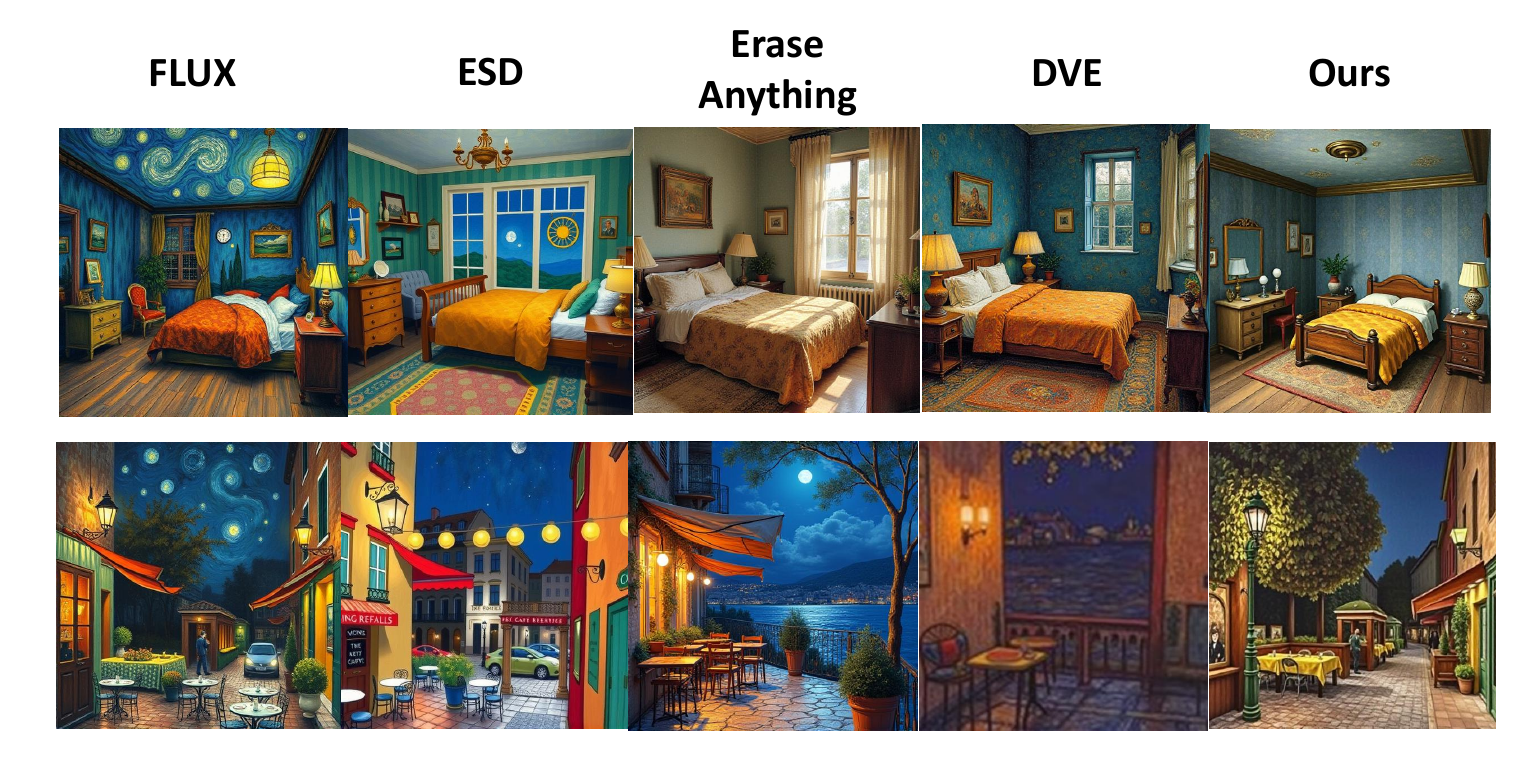}
  \vspace{-0.8em}
  \caption{Comparison of \textbf{Van Gogh} erasure results.}
  \label{fig:exp_artist}
\end{figure*}

\begin{figure*}
  \centering
  \resizebox{\linewidth}{!}{
  \includegraphics{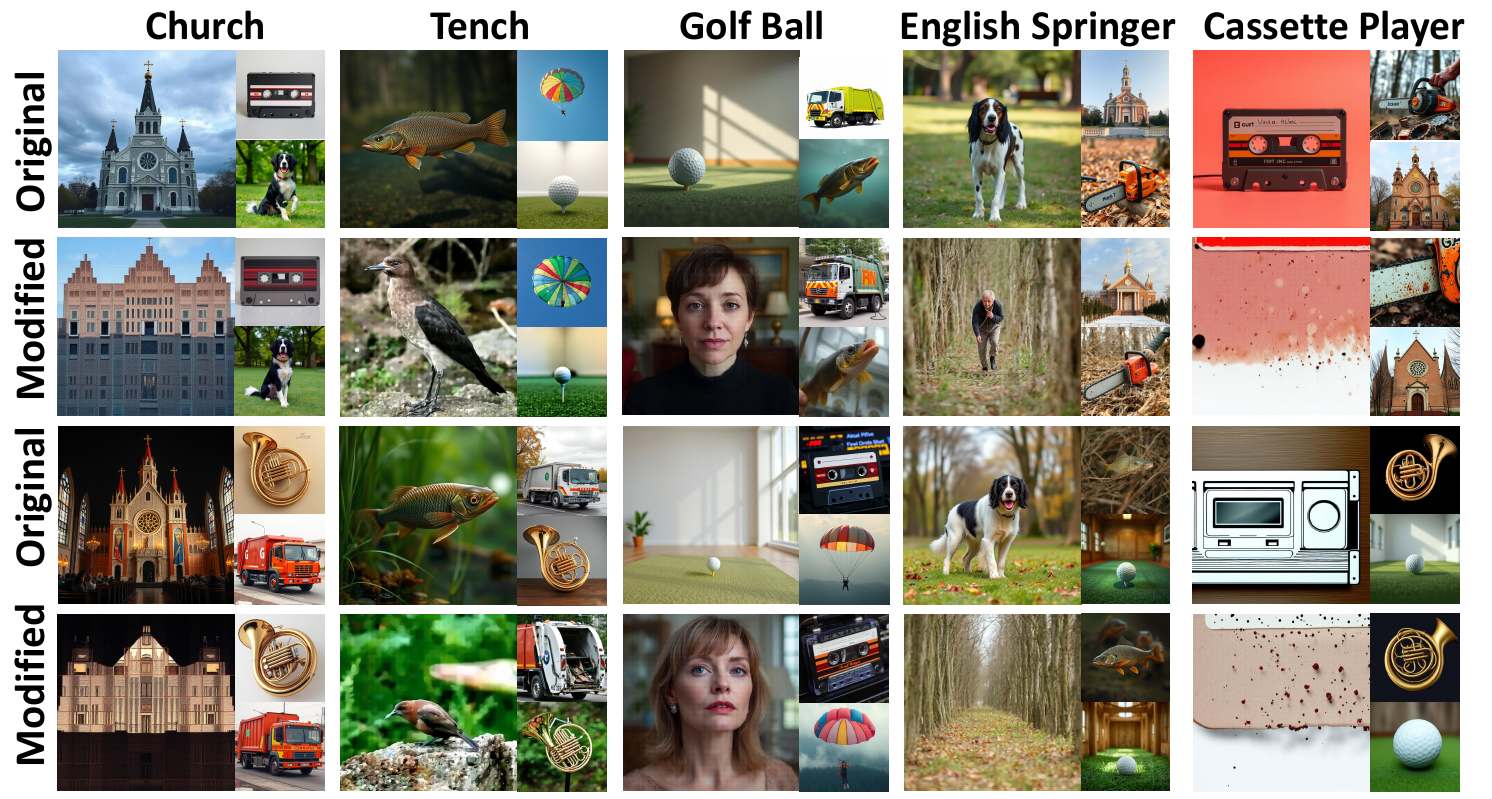}}
  \caption{Additional erasure results of 5 object. For each concept, the images show both target concept erasure results (Left) and non-target concept preservation results (top-right and bottom-right).}
 \label{fig:app_obj_1}
\end{figure*}

\begin{figure*}
  \centering
  \resizebox{\linewidth}{!}{
  \includegraphics{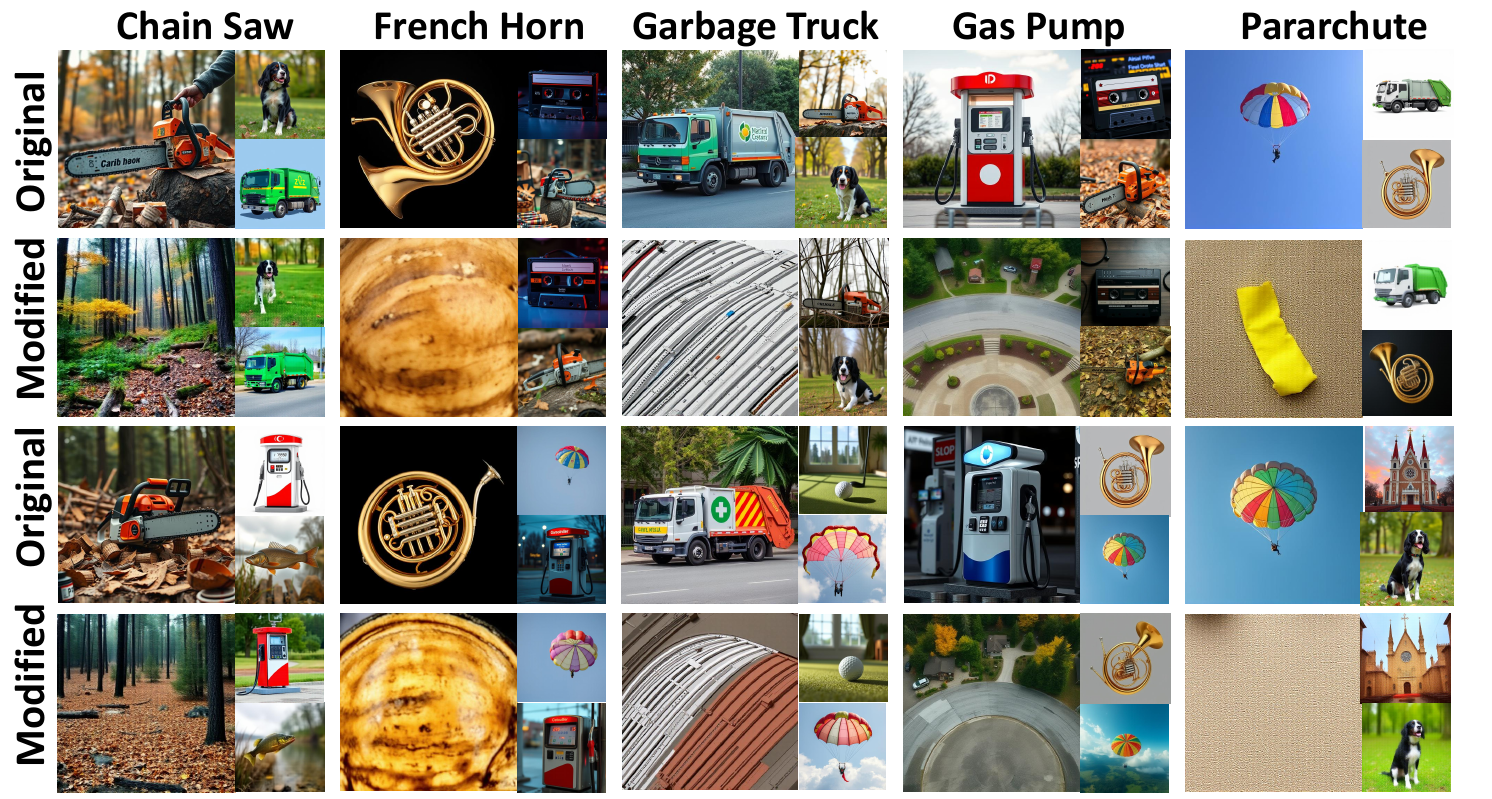}}
  \caption{Additional erasure results of the other 5 object. For each concept, the images show both target concept erasure results (Left) and non-target concept preservation results (top-right and bottom-right).}
 \label{fig:app_obj_2}
\end{figure*}

\begin{figure*}
  \centering
  \resizebox{0.8\linewidth}{!}{
  \includegraphics{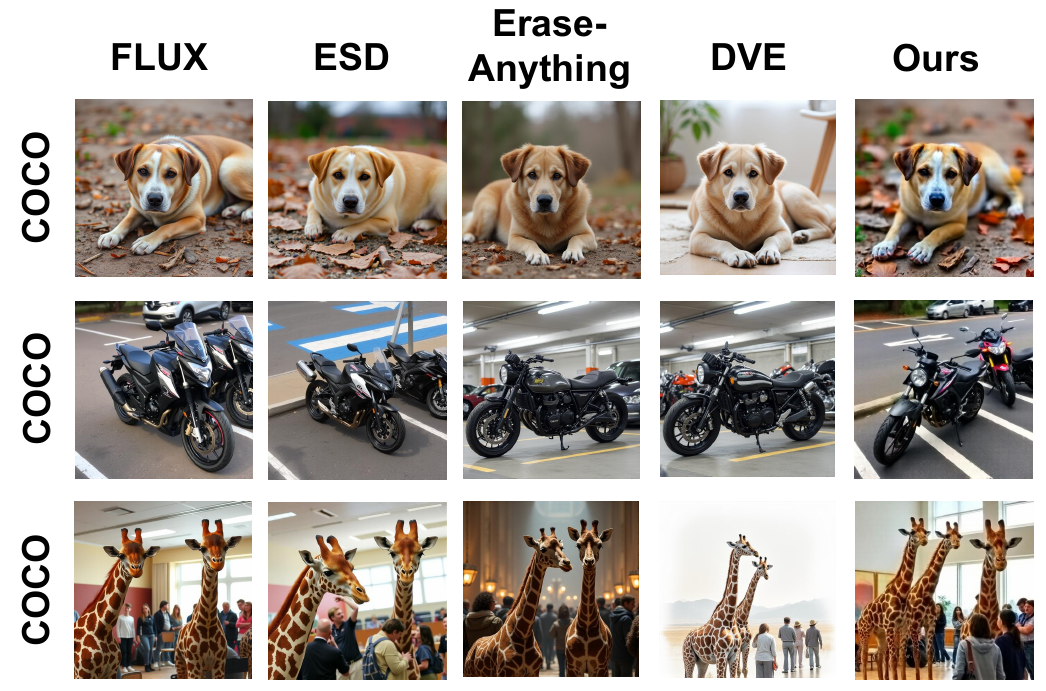}}
  \caption{Comparison of images generated by different methods via MS-COCO dataset.}
 \label{fig:app_coco_1}
\end{figure*}

In this section, we will give more experimental results. In Figure~\ref{fig:app_nudity.pdf} we present additional erasure results generated by prompts from I2P dataset. In Figure~\ref{fig:attacks_1}, we give more results of adversarial attacks, including MMA, RAB,P4D and UnlearnDiff. In Figure~\ref{fig:app_multi}, we compare. the visual results of our multi-concepts erasure tasks. Additional artist style erasure results can be seen in Figure~\ref{fig:app_vangogh} and Figure~\ref{fig:exp_artist}. Additional object erasure results in Figure~\ref{fig:app_obj_1} and Figure~\ref{fig:app_obj_2}.  In Figure~\ref{fig:app_coco_1}, we compare our MSCOCO results with the baseline and we also present the visual results on MS-COCO Dataset for all three types of concept erasure tasks, including "Nudity", "Van Gogh", "Church", etc.

\section{Limitation}
\label{app:limitation}

Our RL-based erasure method demands significant computational resources and training time. Additionally, our method suffers from the limited generation diversity, a common issue in reinforcement learning approaches.

\begin{figure*}[t]
  \centering
  \resizebox{\linewidth}{!}{
  \includegraphics{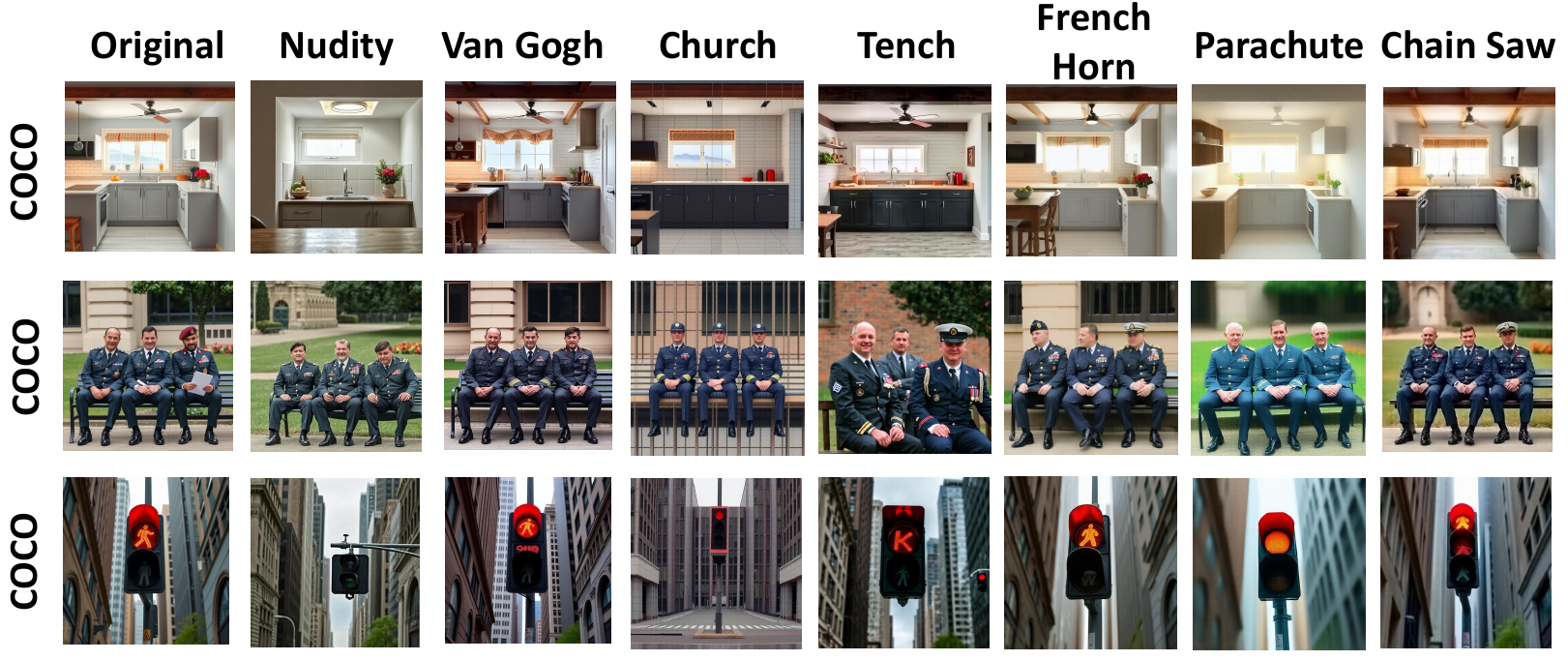}}
  \caption{Comparison of FlowErase-RL on MS-COCO Dataset for all three types of concept erasure tasks, including "Nudity", "Van Gogh", "Church", etc.}
 \label{fig:app_coco_2}
\end{figure*}

\clearpage
\section*{NeurIPS Paper Checklist}

\begin{enumerate}

\item {\bf Claims}
    \item[] Question: Do the main claims made in the abstract and introduction accurately reflect the paper's contributions and scope?
    \item[] Answer: \answerYes{} 
    \item[] Justification: We clearly show our contributions and scope in the abstract and introduction.

\item {\bf Limitations}
    \item[] Question: Does the paper discuss the limitations of the work performed by the authors?
    \item[] Answer: \answerYes{} 
    \item[] Justification: We discuss the limitation of our method in Appendix~\ref{app:limitation}.

\item {\bf Theory assumptions and proofs}
    \item[] Question: For each theoretical result, does the paper provide the full set of assumptions and a complete (and correct) proof?
    \item[] Answer: \answerNA{} 
    \item[] Justification: The paper does not include theoretical results.

    \item {\bf Experimental result reproducibility}
    \item[] Question: Does the paper fully disclose all the information needed to reproduce the main experimental results of the paper to the extent that it affects the main claims and/or conclusions of the paper (regardless of whether the code and data are provided or not)?
    \item[] Answer: \answerYes{} 
    \item[] Justification: The paper conducts tests on mainstream concept erasure tasks, datasets, and mainstream adversarial attack datasets, and lists the main evaluation metrics.

\item {\bf Open access to data and code}
    \item[] Question: Does the paper provide open access to the data and code, with sufficient instructions to faithfully reproduce the main experimental results, as described in supplemental material?
    \item[] Answer: \answerNo{} 
    \item[] Justification: Haven't provide code.

\item {\bf Experimental setting/details}
    \item[] Question: Does the paper specify all the training and test details (e.g., data splits, hyperparameters, how they were chosen, type of optimizer) necessary to understand the results?
    \item[] Answer: \answerYes{} 
    \item[] Justification: We provide detailed statistics and calculation methods for each metric in the paper.

\item {\bf Experiment statistical significance}
    \item[] Question: Does the paper report error bars suitably and correctly defined or other appropriate information about the statistical significance of the experiments?
    \item[] Answer: \answerNo{} 
    \item[] Justification: We demonstrate the statistical significance through extensive experiments on multiple models and various concepts.

\item {\bf Experiments compute resources}
    \item[] Question: For each experiment, does the paper provide sufficient information on the computer resources (type of compute workers, memory, time of execution) needed to reproduce the experiments?
    \item[] Answer: \answerYes{} 
    \item[] Justification: We report our GPU resource in the section of experiment setting.
    
\item {\bf Code of ethics}
    \item[] Question: Does the research conducted in the paper conform, in every respect, with the NeurIPS Code of Ethics \url{https://neurips.cc/public/EthicsGuidelines}?
    \item[] Answer: \answerYes{} 
    \item[] Justification: The research conform with the Ethics.

\item {\bf Broader impacts}
    \item[] Question: Does the paper discuss both potential positive societal impacts and negative societal impacts of the work performed?
    \item[] Answer: \answerYes{} 
    \item[] Justification: Our paper discusses the contributions of our method  to the safety and privacy of high quality image generation.
    
\item {\bf Safeguards}
    \item[] Question: Does the paper describe safeguards that have been put in place for responsible release of data or models that have a high risk for misuse (e.g., pre-trained language models, image generators, or scraped datasets)?
    \item[] Answer: \answerNA{} 
    \item[] Justification: The concept erasure method proposed in our paper is a defensive approach, and the datasets used are publicly available. It does not generate new risky content and therefore does not pose such risks.

\item {\bf Licenses for existing assets}
    \item[] Question: Are the creators or original owners of assets (e.g., code, data, models), used in the paper, properly credited and are the license and terms of use explicitly mentioned and properly respected?
    \item[] Answer: \answerYes{} 
    \item[] Justification: We provide the source and citation for every formula, theorem, model, and dataset.

\item {\bf New assets}
    \item[] Question: Are new assets introduced in the paper well documented and is the documentation provided alongside the assets?
    \item[] Answer: \answerNA{}
    \item[] Justification: The paper does not release new assets.

\item {\bf Crowdsourcing and research with human subjects}
    \item[] Question: For crowdsourcing experiments and research with human subjects, does the paper include the full text of instructions given to participants and screenshots, if applicable, as well as details about compensation (if any)? 
    \item[] Answer: \answerNA{} 
    \item[] Justification: The paper does not involve crowdsourcing nor research with human subjects.

\item {\bf Institutional review board (IRB) approvals or equivalent for research with human subjects}
    \item[] Question: Does the paper describe potential risks incurred by study participants, whether such risks were disclosed to the subjects, and whether Institutional Review Board (IRB) approvals (or an equivalent approval/review based on the requirements of your country or institution) were obtained?
    \item[] Answer: \answerNA{} 
    \item[] Justification: The paper does not involve crowdsourcing nor research with human subjects.

\item {\bf Declaration of LLM usage}
    \item[] Question: Does the paper describe the usage of LLMs if it is an important, original, or non-standard component of the core methods in this research? Note that if the LLM is used only for writing, editing, or formatting purposes and does \emph{not} impact the core methodology, scientific rigor, or originality of the research, declaration is not required.
    \item[] Answer: \answerNA{} 
    \item[] Justification: LLM is used only for writing, editing, and formatting purposes.

\end{enumerate}

\end{document}